\documentclass[10pt,twocolumn,letterpaper]{article}

\usepackage{arxivstyle}   
\newcommand{\tabletextsize}{\scriptsize}
\definecolor{lorablue}{HTML}{EAF2FF}
\definecolor{traingray}{HTML}{F3F4F6}
\definecolor{datagreen}{HTML}{EAF7EE}
\definecolor{trainblue}{HTML}{EAF3FF}
\definecolor{trainborder}{HTML}{4A78A8}
\definecolor{testred}{HTML}{E8E0E8}
\definecolor{testborder}{HTML}{B94A48}
\definecolor{insightbg}{HTML}{FFF7DA}
\definecolor{MorandiMauve}{HTML}{9C829E}
\definecolor{MorandiLavender}{HTML}{AAA7BC}
\definecolor{MorandiNavy}{HTML}{314366}
\definecolor{MorandiSlate}{HTML}{53678F}
\definecolor{MorandiMist}{HTML}{A7AFC2}
\definecolor{MorandiGain}{HTML}{526F61}
\definecolor{MorandiDrop}{HTML}{91616B}
\colorlet{goodgreen}{MorandiGain}
\colorlet{badred}{MorandiDrop}
\colorlet{RewardGray}{MorandiMist!22!white}
\colorlet{RewardBlue}{MorandiSlate!18!white}
\colorlet{RewardYellow}{MorandiLavender!30!white}
\colorlet{RewardGreen}{MorandiSlate!27!white}
\colorlet{RewardGreenText}{MorandiNavy}

\newcommand{\ourmethod}{\textup{\textsc{VideoVera}}\xspace}
\newcommand{\ourdata}{\textup{\textsc{VideoVera}}-27k\xspace}
\newcommand{\ourbenchmark}{\textup{\textsc{VideoVera-Bench}}\xspace}

\newcommand{\papercode}{\href{https://github.com/Yiling-Ma/VideoVera}{github.com/Yiling-Ma/VideoVera}}
\newcommand{\paperdataset}{\href{https://huggingface.co/datasets/YilingMa/VideoVera-6.4k}{VideoVera-6.4k}\;\textbar\;\href{https://huggingface.co/datasets/YilingMa/VideoVera-12k}{VideoVera-12k}\;\textbar\;\href{https://huggingface.co/datasets/YilingMa/VideoVera-27k}{VideoVera-27k}}
\newcommand{\paperproject}{\href{https://yiling-ma.github.io/VideoVera-website/}{yiling-ma.github.io/VideoVera-website/}}
\newcommand{\papercorrespondence}{\texttt{yiling.ma@yale.edu}}

\newcommand{\papertitle}{\resizebox{\linewidth}{!}{\raisebox{-0.12em}{\includegraphics[height=0.95em]{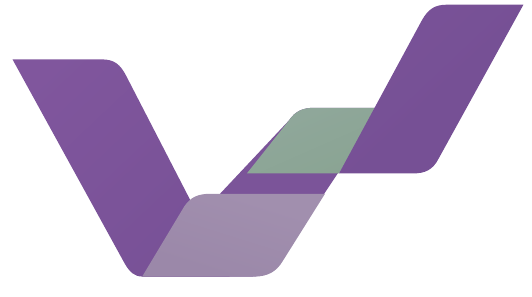}}\hspace{0.3em}\acctitle{\ourmethod:} Verifiable Process Reward for Video Reasoning}}
\newcommand{\paperauthors}{Yiling Ma$^{1}$, Chuhan Li$^{2}$, Xin Eric Wang$^{2}$, Ming-Hsuan Yang$^{3}$}
\newcommand{\paperaffiliations}{$^{1}$Yale University, \ $^{2}$University of California, Santa Barbara, \ $^{3}$University of California, Merced}

\hypersetup{pdftitle={VideoVera: Verifiable Process Reward for Video Reasoning},
            pdfauthor={Yiling Ma, Chuhan Li, Xin Eric Wang, Ming-Hsuan Yang}}

\long\def\paperabstract{%
A correct final answer can conceal errors in a multimodal foundation model's intermediate reasoning, yet final-answer rewards provide no direct feedback on these errors.
We introduce \ourmethod, an annotation-driven pipeline that generates video question-answer pairs and verifiable reasoning traces from existing dataset annotations.
Across eight spatial, temporal, and route reasoning tasks, task-specific rules convert action segments, 3D object locations, camera poses, and navigation trajectories into reference answers and ordered reasoning steps.
These traces enable deterministic checking of model-generated steps and support both process evaluation and graded reinforcement learning rewards, without per-example manual or LLM-based trace annotation.
Using this pipeline, we construct \ourdata for GRPO fine-tuning of Qwen3-VL-8B-Instruct and \ourbenchmark, a benchmark of 1,000 examples with sampled human quality checks.
Our trained model achieves the highest MCQ accuracy (55.8\%) and joint trace-and-answer accuracy (22.9\%) among evaluated models on \ourbenchmark, with transfer gains on selected external spatial and temporal reasoning tasks.
We plan to release the model weights and source code.
}

\begin{document}
\raggedbottom 
\twocolumn[{%
  \arxivtitlecard{\papertitle}{\paperauthors}{\paperaffiliations}{\paperabstract}{%
    \metaline{\faGithub}{Code:}{\papercode}%
    \metaline{\faHuggingFace}{Dataset:}{\paperdataset}%
    \metaline{\faGlobe}{Project:}{\paperproject}%
    \metaline{\faEnvelope[regular]}{Correspondence:}{\papercorrespondence}}
\centering
\includegraphics[width=\textwidth]{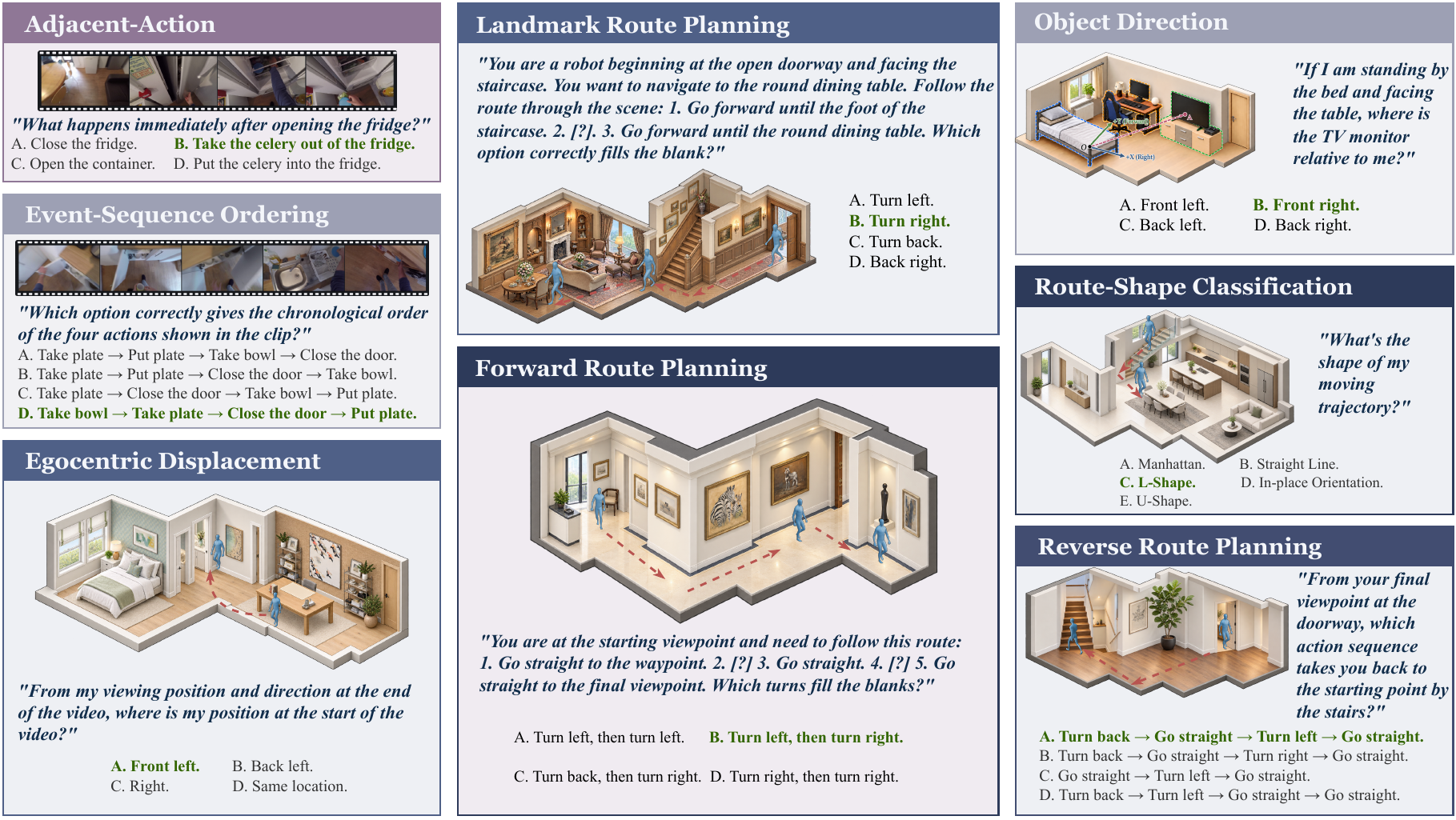}
\captionsetup{type=figure,hypcap=false}
\captionof{figure}{%
  \textbf{Overview of the eight tasks in \ourmethod.}
  Each panel shows one representative multiple-choice question, with the
  correct option highlighted in green.
  Route-Shape Classification additionally visualizes the underlying
  trajectory.}
\label{fig:task_examples}

  \vspace{4pt}
}]

\section{Introduction}
\label{sec:intro}

\begin{figure}[t]
    \centering
    \includegraphics[width=\linewidth]{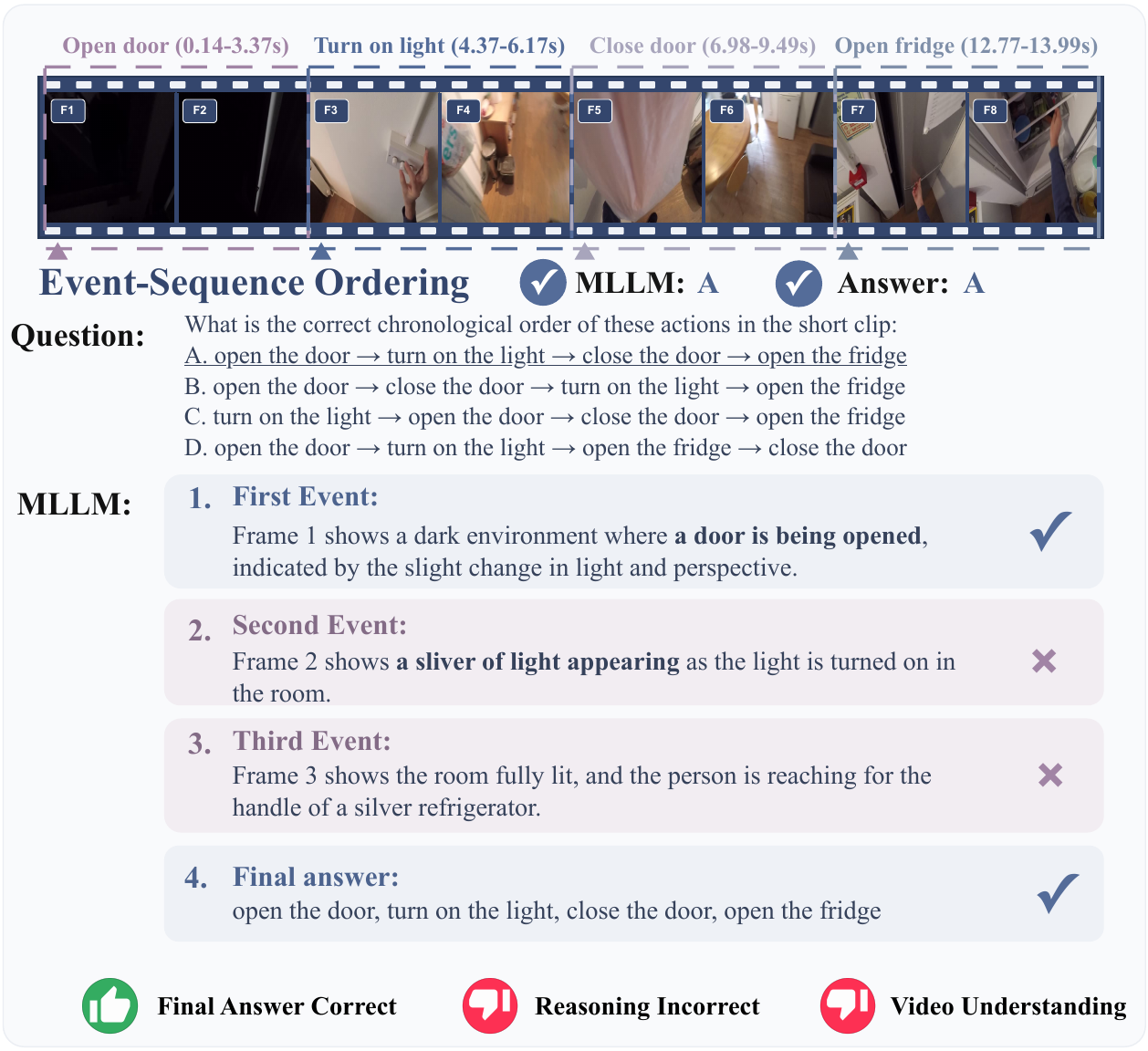}

    \caption{
    \textbf{Correct answers can mask reasoning errors.} 
    The model selects A despite incorrect intermediate steps.
}

    \label{fig:correct_answer_incorrect_trace}
\end{figure}

Multimodal foundation models (MFMs) are increasingly used to help embodied agents understand their surroundings and navigate through physical environments~\citep{cheng2024navila, chen2026egoplan, li2026saw, majumdar2024openeqa}.
Solving these tasks often involves multiple intermediate reasoning steps.
Video question answering provides a way to study the spatial and temporal reasoning capabilities underlying these applications, yet model performance is commonly measured through final-answer accuracy. 
Recent studies~\citep{qi2025vcr, kancheti2026faithful} have shown that a correct final answer can coexist with errors in intermediate reasoning steps, revealing a gap between answer correctness and reasoning correctness.
\Cref{fig:correct_answer_incorrect_trace} illustrates an Event-Sequence Ordering
example in which the model selects the correct option but makes incorrect
intermediate event claims.
Thus, video question-answering datasets annotated only with final answers cannot distinguish these cases from those with correct intermediate reasoning, nor provide direct supervision for the intermediate steps. 

\begin{figure*}[t]
    \centering
    \includegraphics[width=\textwidth]{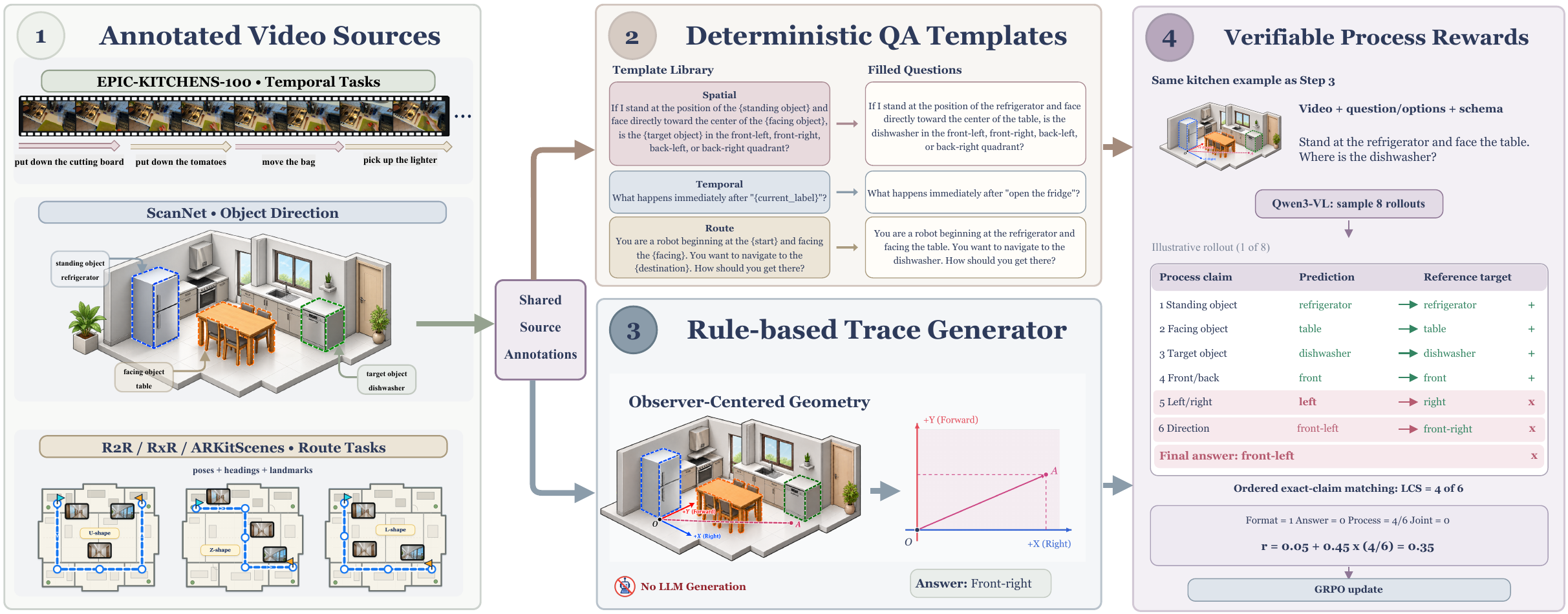}
    \caption{
    \textbf{Overview of the \ourmethod data construction and training
    pipeline.}
    (1) Source annotations are collected from egocentric videos, reconstructed 3D scenes, and navigation trajectories.
    (2) Task-specific templates generate multiple-choice questions and answer choices.
    (3) Rule-based trace generation derives the reference reasoning steps and correct answer from the same source annotations.
    (4) The model predicts reasoning traces and final answers from the video and question. A deterministic verifier compares predictions against reference targets to compute process rewards for GRPO training.
    }
    \label{fig:data_construction}
\end{figure*}

\flushbottom
Addressing this limitation calls for video reasoning data with verifiable supervision for intermediate steps.
However, constructing and checking intermediate targets for individual examples can require substantial annotation effort or additional model inference, which can be costly and difficult to scale. 
Existing work constructs reference step reasoning traces through model-assisted annotation and human verification~\citep{qi2025vcr}, combines LLM-based judgments supplemented with ground-truth bounding box matching~\citep{kancheti2026faithful}, or scores a reasoning prefix by generating multiple possible completions and measuring how often they reach the correct final answer~\citep{wu2026processthinker}. 
These costs and dependencies motivate a data generation approach in which intermediate targets can be computed and checked directly from existing annotations, without verification by external models or humans.

Existing video datasets contain annotations of scene geometry, camera poses, navigation trajectories, and temporal events that provide a basis for constructing such targets ~\citep{dai2017scannet,baruch2021arkitscenes, anderson2018vision,ku2020room,damen2022rescaling}.
Building on top of these datasets, we introduce \ourmethod, a scalable data generation pipeline that converts these annotations into video questions, final answers, and verifiable intermediate steps.
The pipeline parses source annotations into task-specific question-answering templates, as illustrated in \cref{tab:task_type_definitions}.
For each question, the final answer and reference steps are computed from the same underlying annotated state, giving each intermediate target an explicit basis for verification. 
Once the template generation rules are defined, additional examples can be constructed from eligible objects, routes, and event sequences from the datasets directly without separately annotating a reasoning trace for every example. 
A deterministic verifier then compares model-generated structured steps with the reference targets, checking both quality and order. 
The resulting targets support both process evaluation and training with verifiable rewards from intermediate steps.

Using this pipeline, we construct \ourdata, a 27k-example training corpus and \ourbenchmark, a held-out benchmark of 1,000 examples spanning eight spatial, navigation, and temporal reasoning tasks.
\ourbenchmark measures  intermediate-step correctness alongside final-answer accuracy, assessing individual claims, complete traces, and whether both trace and final answer are correct.
Whereas VCR-Bench~\citep{qi2025vcr} uses model-based semantic matching to evaluate reasoning steps against reference steps, \ourbenchmark evaluates predefined structured outputs through deterministic matching against targets derived from source annotations.
This approach provides reproducible evaluation scoring of the specified intermediate claims and their ordering, allowing correct answers with incorrect intermediate steps to be identified separately. 
The training corpus provides supervision in the same structured format, connecting the benchmark's evaluation criteria with directly usable training targets.

To evaluate the effectiveness of this supervision, we train Qwen3-VL-8B~\citep{bai2025qwen3} using a supervised fine-tuning cold start followed by standard GRPO~\citep{shao2024deepseekmath} combining final-answer rewards with rewards for intermediate-step correctness.
We compare against Outcome-only under the same initialization and training recipe to isolate the contribution of verifiable process supervision.
On \ourbenchmark, our model achieves 22.9\% joint trace-and-answer accuracy, compared with 9.3\% for the base model and 22.2\% for
Outcome-only.
MCQ accuracy also increases to 55.8\%, compared with 39.9\% and 52.3\%, respectively.
Results on four external benchmarks, including ReVSI, SAW-Bench, TOMATO,
and MVBench, further show improvements over the base model on spatial reasoning, temporal reasoning, and video understanding tasks.

Our contributions are summarized as below: 
\begin{itemize}
    \item We introduce \ourmethod, a scalable pipeline that generates video QA pairs and verifiable reasoning traces from existing annotations for process supervision.
    \item We introduce \ourbenchmark to deterministically evaluate reasoning steps, their order, and final answers.
    \item We demonstrate improvements in answer and trace correctness through verifiable process supervision, with gains on out-of-domain video benchmarks.
\end{itemize}

\section{Related Work}
\paragraph{Video understanding benchmarks.}
Existing video understanding benchmarks evaluate MFMs on general-purpose video question answering~\citep{li2024mvbench, fu2025video}, long-video comprehension~\citep{chandrasegaran2024hourvideo, wang2025lvbench, wu2024longvideobench}, visual spatial reasoning~\citep{yang2025thinking, zhang2026revsi, li2026saw, ju2026momagraph, sibench}, visual temporal reasoning~\citep{shangguan2025tomato, liu2024tempcompass, chen2024rextime}, and domain-specific reasoning~\citep{zhao2025mmvu, he2024mmworld, yadav2026videonet}.
These benchmarks broaden the range of capabilities being assessed, but task-level performance alone provides limited insight into the correctness of intermediate reasoning steps.
Recent work has therefore extended evaluation beyond final-answer correctness to assess intermediate reasoning~\citep{qi2025vcr, nagrani2025minerva, nagrani2026minervaego}.
However, these benchmarks require human annotation or verification of question-specific reference reasoning traces, incurring additional annotation costs as they expand.
To reduce this annotation burden, we introduce \ourmethod, a scalable data generation pipeline that converts structured annotations in existing video datasets into questions, answers, and verifiable intermediate reasoning targets.

\paragraph{Video reasoning methods.}
Methods for improving video reasoning have explored supervised learning on grounded reasoning traces and reinforcement learning.
Supervised approaches connect reasoning to visual evidence through annotations of relevant frames and objects~\citep{han2025videoespresso} or explicit frame references within reasoning traces~\citep{ghazanfari2026chain}.
Reinforcement learning further develops spatial and temporal reasoning through temporal reward shaping~\citep{feng2026video} and training that combines forward and reverse route reasoning~\citep{wu2026st}.
To guide reasoning beyond final-answer supervision, existing methods structure outputs into summarization, thinking, and answer stages~\citep{wang2026reinforcing}, separate perception from reasoning~\citep{jiang2026videop2r}, or estimate step utility from sampled continuations~\citep{wu2026processthinker}.
Other approaches learn reward models from response preferences~\citep{wei2026video} or incorporate logical consistency and visual grounding constraints into policy optimization for visual spatial reasoning~\citep{kancheti2026faithful}.
More directly related to our work, VideoTrace-R1~\citep{caovideotrace} computes deterministic rewards for intermediate temporal claims using structured temporal traces.
Our work focuses on generating questions, answers, and verifiable intermediate targets from existing annotations in real-world video datasets, providing scalable supervision for both outcome and process rewards.

\section{\ourmethod}
\label{sec:method}
\definecolor{AlgorithmHighlight}{HTML}{D8DEE9}
\newcommand{\algkey}[1]{%
  {\setlength{\fboxsep}{1.5pt}%
   \colorbox{AlgorithmHighlight}{\strut\textbf{#1}}}}

\begin{algorithm}[t]
\caption{Training \ourmethod}
\label{alg:videoprm}
\small
\begin{algorithmic}[1]
\Require Policy-view data $\mathcal D_{\mathrm{pol}}=\{(i,x_i)\}$
\Require Hidden targets $\mathcal H[i]=(G_i,a_i^*)$
\Require Cold-start policy $\pi_0$ and group size $g=8$

\State Initialize $\pi_\theta\gets\pi_0$
\State Set frozen reference $\pi_{\mathrm{ref}}\gets\pi_0$

\For{each $(i,x_i)$ in the fixed training order}
    \State Snapshot $\pi_{\mathrm{old}}\gets\pi_\theta$
    \State Sample $\{y_j\}_{j=1}^{g}\sim\pi_{\mathrm{old}}(\cdot\mid x_i)$
    with task constraints
    \State \algkey{Retrieve hidden targets}
    $(G_i,a_i^*)\gets\mathcal H[i]$

    \For{$j=1,\ldots,g$}
        \If{$y_j$ fails parsing or schema validation}
            \State $r_j\gets0$
        \Else
            \State Parse $(\hat G_j,\hat a_j)$ and compute
            $R_{p,j}$
            \State \algkey{Compute reward}
            $r_j$ via \eqref{eq:structured_reward} using $(G_i,a_i^*)$
        \EndIf
    \EndFor

    \If{$\max_j r_j-\min_j r_j>10^{-8}$}
        \State For all $j$, compute
        $A_j\gets
        \frac{r_j-\operatorname{mean}(r)}
        {\operatorname{std}_{\mathrm{pop}}(r)+10^{-4}}$
        \State Cache token log-probabilities under
        $\pi_{\mathrm{old}}$ and $\pi_{\mathrm{ref}}$
        \State Run two clipped GRPO epochs with KL to $\pi_{\mathrm{ref}}$
    \EndIf
\EndFor
\end{algorithmic}
\end{algorithm}
We detail the construction of verifiable reasoning traces and their use in process verification and reward computation.

\subsection{Problem Formulation}
\label{sec:formulation}
For a task type $T$, the model receives an input $x=(v,q)$, where $v$ is a video and $q$ is a multiple-choice question with options.
A rule-based trace generator constructs a reference reasoning
trace $G$ from source annotations, such as object locations,
camera poses, navigation trajectories, or action timestamps.
Each example has a correct answer $a$ and an ordered reference
trace $G=(c_1,\ldots,c_M)$, where $c_i$ denotes the $i$-th
required reasoning step for $i=1,\ldots,M$.
During generation, the model receives only $x$ and the output
instructions for task $T$, and predicts a reasoning trace
$\hat{G}=(\hat{c}_1,\ldots,\hat{c}_N)$ and a final answer $\hat{a}$.
After generation, the deterministic verifier compares
$\hat{G}$ with $G$ in content and order and checks whether
$\hat{a}=a$.
\paragraph{Scope of process verification.}
Our verifier checks annotation-grounded intermediate claims and their ordering, measuring output-level process correctness rather than causal reasoning faithfulness. It does not establish whether the generated steps reflect the model's internal reasoning.

\subsection{Constructing Verifiable Process Data}
\label{sec:data_construction}
For each example, the source annotations determine both the required reasoning process and the final answer.
As illustrated in \cref{fig:data_construction}, the pipeline (1) selects a source example (\textit{e.g.}, from ScanNet or EPIC-KITCHENS-100), (2) generates
questions and answers, (3) constructs reference traces
through rule-based generation, and (4) verifies the predicted reasoning steps for process evaluation
and reward computation.
During generation, the model receives only the video, question, answer choices, and output instructions.
The pipeline covers eight spatial, temporal, and route reasoning tasks. \Cref{fig:task_examples} illustrates examples, and \cref{tab:task_type_definitions} summarizes task definitions and required reasoning steps.

\definecolor{TaskSpatialBlue}{HTML}{E1E7EF}
\definecolor{TaskRouteMint}{HTML}{E4EAE2}
\definecolor{TaskTemporalPeach}{HTML}{F0EBEF}

\begin{table*}[t]
\centering
\caption{
\textbf{Task definitions, sources, and required process claims
for eight tasks.}
Shading distinguishes
\colorbox{TaskSpatialBlue}{\vphantom{route}\smash{spatial}},
\colorbox{TaskRouteMint}{route}, and
\colorbox{TaskTemporalPeach}{\vphantom{route}\smash{temporal}} tasks.
EPIC-100 denotes EPIC-KITCHENS-100.
Matterport3D videos are rendered using routes derived from
R2R~\cite{anderson2018vision} and RxR~\cite{ku2020room}.
}
\label{tab:task_type_definitions}
\tabletextsize
\setlength{\tabcolsep}{4pt}
\renewcommand{\arraystretch}{1.10}

\begin{tabularx}{\textwidth}{
    >{\raggedright\arraybackslash}p{0.165\textwidth}
    >{\raggedright\arraybackslash}p{0.115\textwidth}
    >{\raggedright\arraybackslash}p{0.28\textwidth}
    >{\raggedright\arraybackslash}X
}
\toprule
\textbf{Task Type}
& \textbf{Source}
& \textbf{Definition}
& \textbf{Required Process Claims} \\
\midrule

\cellcolor{TaskSpatialBlue}\textbf{Object Direction}
& ScanNet~\cite{dai2017scannet};
  ARKitScenes~\cite{baruch2021arkitscenes}
& Target direction in an observer-centered frame.
& Observer, facing landmark, target, front/back relation,
left/right relation, final direction. \\

\cellcolor{TaskSpatialBlue}\textbf{Egocentric Displacement}
& Matterport3D~\cite{chang2017matterport3d}
& One trajectory endpoint relative to the view at the other endpoint.
& Reference/target endpoints, ordered movements,
final egocentric direction. \\

\midrule

\cellcolor{TaskRouteMint}\textbf{Forward Route Planning}
& Matterport3D~\cite{chang2017matterport3d}
& Navigation actions from the starting pose to the destination.
& Indexed forward movements, indexed turns,
complete forward route. \\

\cellcolor{TaskRouteMint}\textbf{Landmark Route Planning}
& ARKitScenes~\cite{baruch2021arkitscenes}
& Missing turns along a landmark-defined route.
& Starting landmark, facing landmark,
route-leg destinations, ordered turns. \\

\cellcolor{TaskRouteMint}\textbf{Route-Shape Classification}
& Matterport3D~\cite{chang2017matterport3d}
& Geometric shape of the trajectory.
& Ordered motions/turns, final shape. \\

\cellcolor{TaskRouteMint}\textbf{Reverse Route Planning}
& Matterport3D~\cite{chang2017matterport3d}
& Navigation actions from the final pose to the start.
& Forward route, inverse actions, reverse temporal order. \\

\midrule

\cellcolor{TaskTemporalPeach}\textbf{Event-Sequence Ordering}
& EPIC-100~\cite{damen2022rescaling}
& Chronological order of four events.
& Events, temporal positions, order. \\

\cellcolor{TaskTemporalPeach}\textbf{Adjacent-Action}
& EPIC-100~\cite{damen2022rescaling}
& Action before or after an anchor.
& Anchor, adjacent action, before/after relation. \\

\bottomrule
\end{tabularx}
\end{table*}

\paragraph{Data sources.}
We use annotations from EPIC-KITCHENS-100~\cite{damen2022rescaling}, ScanNet~\cite{dai2017scannet}, and
ARKitScenes~\cite{baruch2021arkitscenes}, together with navigation trajectories from R2R~\cite{anderson2018vision} and
RxR~\cite{ku2020room}, both defined over Matterport3D
environments~\cite{chang2017matterport3d}. 
Together, these sources provide timestamped action segments, 3D object locations, camera poses, and ordered
navigation paths.

\paragraph{Question-answer generation.}
We automatically generate each question and its answer choices from source annotations using task-specific templates (\cref{fig:data_construction}, (2)). 
Each template specifies the queried relation, question wording, allowed answers, and distractor construction.

\paragraph{Task-specific trace generation.}
The rule-based trace generator follows the task definitions in \cref{tab:task_type_definitions}.
For spatial tasks, the generator computes horizontal directions using the observer's position and facing landmark for object direction or the camera pose at the reference endpoint for egocentric displacement.
For route tasks, the generator extracts movements, turns, landmarks, and shapes from annotated trajectories, including reversed routes, to verify trajectory reconstruction rather than arbitrary valid paths.
For temporal tasks, the generator orders annotated action segments or identifies the action immediately before or after an anchor. 

\paragraph{Training data and benchmark.}
The pipeline produces \ourdata and \ourbenchmark, which contains 1,000 examples (125 per task).
Benchmark examples are disjoint from the training corpus by example identifier, source media group, and decoded video content.

\subsection{Human Quality Control}
\label{sec:human_quality_control}
To assess data quality, we select 800 examples from the training corpus and benchmark through stratified sampling across eight tasks.
Two reviewers independently watch each video without seeing the generated answer or reference trace. 
They assess whether the question is answerable from the video, select an answer option, and record the reasoning steps.
We compare their annotations with the generated answer and reference trace to identify incorrect answers, unsupported steps, or ordering errors.
The review yields 97.9\% answer agreement, 97.4\% trace
validity, and 97.4\% joint example acceptance, with Cohen's $\kappa=0.881$~\citep{cohen1960coefficient} for joint acceptance before adjudication. 

\subsection{Process Verification and Reward}
\label{sec:process_reward}
\paragraph{Trace verification.}
For each output that satisfies the task-specific schema,
including required fields and allowed value types, the verifier standardizes capitalization, whitespace, and trailing periods in text labels before comparing the predicted and reference traces.
A predicted step matches a reference step only when they represent the same reasoning operation and contain the same task-specific content. 
Let $K$ denote the length of their longest common subsequence, with predicted and reference trace lengths $N$ and $M$. The process reward is $R_{p}=\frac{2K}{N+M}$.
The reward equals one only for an exact ordered trace and decreases for missing, additional, incorrect, or reordered claims.

\paragraph{Reward function.}
We use binary indicators $R_f$ for schema validity~\citep{jiang2026videop2r}, $R_a=\mathbf{1}[\hat{a}=a]$ for answer correctness, and $R_j=\mathbf{1}[\hat{a}=a \land \hat{G}=G]$ for joint answer and trace correctness. 
The structured reward is
\begin{equation}
\label{eq:structured_reward}
R=\lambda_f R_f+\lambda_a R_a+\lambda_p R_p+\lambda_j R_j.
\end{equation}
Outputs that violate the schema receive zero reward.
We select the weights on a task-balanced validation set without accessing the frozen benchmark, maximizing joint accuracy while preserving answer accuracy and format validity.
We use $(\lambda_f,\lambda_a,\lambda_p,\lambda_j)=(0.05,0.40,0.45,0.10)$. Outcome-only uses $R_a$ alone under the same schema validation.

\paragraph{Policy training.}
We initialize the trainable policy and the
fixed reference policy for KL regularization from the SFT cold-start checkpoint trained on 1,600 reserved examples. 
For each RL prompt, we sample eight schema-constrained responses, then score them against hidden verifier targets retrieved by example ID. 
For groups with varying rewards, we compute group-normalized advantages and run two epochs of clipped GRPO
updates on the same responses. 
Groups with constant rewards count toward the rollout budget but receive no updates. \Cref{alg:videoprm} summarizes the procedure.

\section{Experiments}
\label{sec:experiments}

\newcommand{\vprgain}[1]{\textsuperscript{\textcolor{goodgreen}{\tiny #1}}}
\newcommand{\vprdrop}[1]{\textsuperscript{\textcolor{badred}{\tiny #1}}}
\newcommand{\vprflat}[1]{\textsuperscript{\textcolor{black!50}{\tiny #1}}}
\definecolor{AvgHighlight}{HTML}{EDF0F5}
\providecommand{\avgcell}[1]{\cellcolor{AvgHighlight}#1}

\definecolor{JointHighlight}{HTML}{EDF0F5}
\providecommand{\jointcell}[1]{\cellcolor{JointHighlight}#1}

\begin{table*}[t]
\centering
\caption{
\textbf{Results on \ourbenchmark.}
All models use 24 uniformly sampled frames, greedy decoding
with task-specific schema constraints, and a 640-token output limit.
%
%
\textbf{Task abbreviations}: OD (Object Direction),
ED (Egocentric Displacement), FRP (Forward Route Planning),
LRP (Landmark Route Planning), RSC (Route-Shape Classification),
RRP (Reverse Route Planning), ESO (Event-Sequence Ordering),
and AA (Adjacent-Action).
\textbf{Bold}: best; \underline{underline}: second best.
}
\label{tab:main_benchmark_results}
\tabletextsize
\setlength{\tabcolsep}{4.0pt}
\renewcommand{\arraystretch}{1.10}

\begin{adjustbox}{max width=\textwidth}
\begin{tabular}{@{}lcccccccccccc@{}}
\toprule
\textbf{Model}
& \multicolumn{4}{c}{\textbf{Aggregate Metrics}}
& \multicolumn{8}{c}{\textbf{Task-wise MCQ Accuracy}} \\
\cmidrule(lr){2-5}
\cmidrule(l){6-13}
& \textbf{Joint}
& \textbf{Claim Rec.}
& \textbf{Exact}
& \textbf{Avg.}
& \textbf{OD}
& \textbf{ED}
& \textbf{FRP}
& \textbf{LRP}
& \textbf{RSC}
& \textbf{RRP}
& \textbf{ESO}
& \textbf{AA} \\
\midrule

\rowcolor{RewardGray}[0pt][0pt]
\multicolumn{13}{@{}l@{}}{\textit{Proprietary MFMs}} \\
\addlinespace[2pt]

Gemini 3.1 Pro~\cite{google2026gemini31pro}
& \jointcell{19.1} & 53.0 & 19.1 & 47.8
& 51.2 & 35.2 & 48.8 & 29.6 & 36.0 & 48.0 & 61.6 & 72.0 \\

\addlinespace[2pt]
\rowcolor{RewardBlue}[0pt][0pt]
\multicolumn{13}{@{}l@{}}{\textit{General-purpose MFMs}} \\
\addlinespace[2pt]

LLaVA-OneVision-1.5-8B-Instruct~\cite{an2025llava}
& \jointcell{4.3} & 39.3 & 5.2 & 30.2
& 18.4 & 22.4 & 40.0 & 23.2 & 21.6 & 25.6 & 44.8 & 45.6 \\

Qwen2.5-VL-7B-Instruct~\cite{bai2025qwen25vltechnicalreport}
& \jointcell{5.3} & 38.2 & 6.5 & 32.8
& 18.4 & 33.6 & 41.6 & 29.6 & 27.2 & 26.4 & 44.8 & 40.8 \\

VideoLLaMA3-7B~\cite{zhang2025videollama}
& \jointcell{6.9} & 49.8 & 8.1 & 34.7
& 17.6 & 32.0 & 27.2 & 40.8 & 22.4 & 38.4 & 55.2 & 44.0 \\

InternVL3.5-8B-Instruct~\cite{wang2025internvl3}
& \jointcell{11.6} & 49.3 & 13.6 & 38.1
& 30.4 & 30.4 & 35.2 & 25.6 & 31.2 & 28.8 & 64.8 & 58.4 \\

Qwen3-VL-8B-Thinking~\cite{bai2025qwen3}
& \jointcell{14.5} & 49.4 & 14.7 & 38.3
& 36.0 & 28.8 & 40.8 & 28.8 & 25.6 & 32.0 & 55.2 & 59.2 \\

\addlinespace[2pt]
\rowcolor{testred}[0pt][0pt]
\multicolumn{13}{@{}l@{}}{\textit{Video reasoning MFMs}} \\
\addlinespace[2pt]

VideoChat-R1-Thinking-7B~\cite{li2025videochat}
& \jointcell{5.0} & 40.5 & 6.1 & 32.7
& 16.8 & 34.4 & 39.2 & 27.2 & 24.8 & 28.8 & 44.0 & 46.4 \\

VideoRFT-7B~\cite{wang2026videorft}
& \jointcell{5.1} & 43.8 & 7.1 & 32.6
& 19.2 & 28.8 & 42.4 & 36.0 & 20.0 & 27.2 & 48.8 & 38.4 \\

Time-R1-7B~\cite{wang2026time}
& \jointcell{5.5} & 39.4 & 6.4 & 32.6
& 16.8 & 28.8 & 41.6 & 22.4 & 25.6 & 33.6 & 49.6 & 42.4 \\

Video-RTS-7B~\cite{videorts}
& \jointcell{6.0} & 38.0 & 7.1 & 33.7
& 21.6 & 32.8 & 43.2 & 30.4 & 25.6 & 31.2 & 45.6 & 39.2 \\

VideoChat-R1-7B~\cite{li2025videochat}
& \jointcell{6.3} & 37.0 & 7.3 & 32.8
& 16.8 & 36.0 & 42.4 & 23.2 & 22.4 & 30.4 & 50.4 & 40.8 \\

Video-R1-7B~\cite{feng2026video}
& \jointcell{7.7} & 45.0 & 8.8 & 36.4
& 28.8 & 32.0 & 44.8 & 31.2 & 32.8 & 28.8 & 49.6 & 43.2 \\

Video-Thinker-7B~\cite{wang2025video}
& \jointcell{8.1} & 39.3 & 8.8 & 33.2
& 20.8 & 24.0 & 28.0 & 39.2 & 30.4 & 40.0 & 46.4 & 36.8 \\

VideoTrace-R1-7B~\cite{caovideotrace}
& \jointcell{8.1} & 38.6 & 8.7 & 34.1
& 20.8 & 24.0 & 46.4 & 28.8 & 28.8 & 40.0 & 47.2 & 36.8 \\

\midrule
\rowcolor{RewardGreen}[0pt][0pt]
\multicolumn{13}{@{}l@{}}{\textit{Base model and fine-tuned variants}} \\
\addlinespace[2pt]
Qwen3-VL-8B-Instruct~\cite{bai2025qwen3}
& \jointcell{9.3} & 40.9 & 9.5 & 39.9
& 36.8 & 34.4 & 49.6 & 20.8 & 31.2 & 31.2 & 56.8 & 58.4 \\

\quad $+$ SFT
& \jointcell{14.1\vprgain{+4.8}}
& 51.8\vprgain{+10.9}
& 15.0\vprgain{+5.5}
& 47.7\vprgain{+7.8}
& \underline{54.4}\vprgain{+17.6}
& \underline{44.0}\vprgain{+9.6}
& 37.6\vprdrop{-12.0}
& 35.2\vprgain{+14.4}
& 24.0\vprdrop{-7.2}
& 45.6\vprgain{+14.4}
& \underline{67.2}\vprgain{+10.4}
& 73.6\vprgain{+15.2} \\

\quad $+$ Outcome-only
& \jointcell{\underline{22.2}\vprgain{+12.9}}
& \underline{61.3}\vprgain{+20.4}
& \underline{23.0}\vprgain{+13.5}
& \underline{52.3}\vprgain{+12.4}
& 48.0\vprgain{+11.2}
& 42.4\vprgain{+8.0}
& \textbf{51.2}\vprgain{+1.6}
& \underline{45.6}\vprgain{+24.8}
& \underline{40.0}\vprgain{+8.8}
& \underline{48.8}\vprgain{+17.6}
& \underline{67.2}\vprgain{+10.4}
& \underline{75.2}\vprgain{+16.8} \\

\quad $+$ \textbf{\ourmethod}
& \jointcell{\textbf{22.9}\vprgain{+13.6}}
& \textbf{63.7}\vprgain{+22.8}
& \textbf{24.5}\vprgain{+15.0}
& \textbf{55.8}\vprgain{+15.9}
& \textbf{58.4}\vprgain{+21.6}
& \textbf{46.4}\vprgain{+12.0}
& \underline{50.4}\vprgain{+0.8}
& \textbf{47.2}\vprgain{+26.4}
& \textbf{42.4}\vprgain{+11.2}
& \textbf{55.2}\vprgain{+24.0}
& \textbf{68.8}\vprgain{+12.0}
& \textbf{77.6}\vprgain{+19.2} \\

\bottomrule
\end{tabular}
\end{adjustbox}
\end{table*}

\subsection{Evaluation Protocol}
\label{sec:evaluation_protocol}

\paragraph{Models.}
We evaluate a diverse set of general-purpose and video reasoning MFMs in a zero-shot setting. 
Open-source general-purpose models include
Qwen2.5-VL~\citep{bai2025qwen25vltechnicalreport},
Qwen3-VL~\citep{bai2025qwen3},
LLaVA-OneVision~\citep{an2025llava},
VideoLLaMA3~\citep{zhang2025videollama},
and InternVL3.5~\citep{wang2025internvl3}.
We also evaluate the proprietary Gemini 3.1~\citep{google2026gemini31pro}
and video reasoning MFMs~\citep{wang2026videorft,
wang2026time,li2025videochat,videorts,
caovideotrace,feng2026video,wang2025video}.

\paragraph{Benchmarks.}
We evaluate models on \ourbenchmark{} and four out-of-domain video understanding benchmarks: ReVSI \citep{zhang2026revsi}, SAW-Bench \citep{li2026saw}, TOMATO \citep{shangguan2025tomato}, and MVBench \citep{li2024mvbench}.
ReVSI and SAW-Bench focus on spatial reasoning, while TOMATO assesses temporal reasoning and MVBench evaluates general video understanding.
We follow each benchmark's official evaluation
protocol whenever possible.

\paragraph{Metrics.}

On \ourbenchmark, we evaluate the eight tasks defined in \cref{tab:task_type_definitions} using both answer-level and process-level metrics. 
Multiple-choice question (MCQ) accuracy measures the fraction of examples with the correct final answer. 
We report MCQ accuracy for each task and the mean across tasks (\textbf{Avg.}).
Process claim recall (\textbf{Claim Rec.}) measures the fraction of reference reasoning steps correctly matched by the predictions. Each step is counted at most once, and results are aggregated across all reference steps. 
Exact trace accuracy (\textbf{Exact}) measures the fraction of examples whose predicted traces contain all required steps in order, with no missing, additional, or
duplicated steps.
Joint accuracy (\textbf{Joint}) measures the fraction of examples with both an exact trace and the correct final answer. 
For out-of-domain benchmarks, we retain their original official metrics and aggregation rules: MCQ accuracy for
SAW-Bench~\citep{li2026saw}, TOMATO~\citep{shangguan2025tomato}, and MVBench~\citep{li2024mvbench}; and exact-match
MCQ accuracy and mean related accuracy (MRA)
for ReVSI's~\citep{zhang2026revsi} numerical questions.

\subsection{Training Setup}
\label{sec:training_setup}

We use Qwen3-VL-8B-Instruct \citep{bai2025qwen3} as the base MFM and train three variants:
(1) \textbf{SFT}, trained on the full training set using supervised fine-tuning;
(2) \textbf{Outcome-only}, optimized with GRPO using binary rewards based solely on final-answer correctness; and
(3) \textbf{\ourmethod}, optimized with GRPO using process rewards (\cref{sec:process_reward}) that evaluate intermediate reasoning steps.
Both RL variants are initialized from the same SFT cold-start checkpoint trained on 1.6k examples and subsequently optimized over a single pass through the remaining 25k examples.

We fine-tune all variants using LoRA \citep{hu2021lora} with $r=16$ and $\alpha=32$.
Both full-data SFT and SFT cold-start training run for one epoch with a learning rate of $1\times10^{-4}$ and a dropout rate of $0.05$.
For RL, we use a learning rate of $5\times10^{-6}$ with 50 learning-rate warm-up steps and disable dropout.
For rollout generation, we use 24 frames per video and schema-constrained sampling with a temperature of $0.8$, top-$p$ of $0.95$, and a maximum generation length of 640 tokens.
GRPO uses a clipping parameter of $0.2$ and a KL regularization coefficient of $0.02$.
All training runs are conducted on eight NVIDIA B200 GPUs.

\subsection{Main Results}
\label{sec:main_results}
\providecommand{\extgain}[1]{\textsuperscript{\textcolor{goodgreen}{\tiny #1}}}
\providecommand{\extdrop}[1]{\textsuperscript{\textcolor{badred}{\tiny #1}}}
\providecommand{\extflat}[1]{\textsuperscript{\textcolor{black!50}{\tiny #1}}}

\begin{table*}[t]
\centering
\caption{
\textbf{Transfer to external video understanding benchmarks.}
We report overall scores and subtasks related to \ourbenchmark's task families: direction, route shape, reverse planning, and temporal order.
\textbf{Bold}: best; \underline{underline}: second best.
}
\label{tab:external_benchmark_results}

\scriptsize
\setlength{\tabcolsep}{2pt}

\begin{adjustbox}{max width=\textwidth}
\begin{tabular}{@{}l*{12}{c}@{}}
\toprule
\textbf{Model}
& \multicolumn{2}{c}{\textbf{ReVSI}}
& \multicolumn{4}{c}{\textbf{SAW-Bench}}
& \multicolumn{3}{c}{\textbf{MVBench}}
& \multicolumn{3}{c}{\textbf{TOMATO}} \\
\cmidrule(lr){2-3}
\cmidrule(lr){4-7}
\cmidrule(lr){8-10}
\cmidrule(l){11-13}

& \textbf{Overall}
& \mbox{Relative Direction}
& \textbf{Overall}
& Direction
& Shape
& \mbox{Reverse Planning}
& \textbf{Overall}
& \mbox{Action Antonym}
& \mbox{Scene Transition}
& \textbf{Overall}
& Direction
& \mbox{Shape \& Trend} \\

\midrule

Base
& \textbf{43.6} & 35.9
& \underline{42.8} & \underline{41.7} & 49.6 & 25.3
& 68.7 & 81.5 & 91.0
& 34.8 & 45.4 & 38.6 \\

\midrule

SFT
& 40.6\extdrop{$-$3.0} & \underline{38.4}\extgain{+2.5}
& 41.8\extdrop{$-$1.0} & 41.4\extdrop{$-$0.3} & 45.8\extdrop{$-$3.8} & \underline{28.4}\extgain{+3.1}
& 68.5\extdrop{$-$0.2} & \underline{83.5}\extgain{+2.0} & \underline{93.5}\extgain{+2.5}
& 32.7\extdrop{$-$2.1} & 46.9\extgain{+1.5} & 35.9\extdrop{$-$2.7} \\

Outcome-only
& \textbf{43.6}\extflat{0.0} & 37.4\extgain{+1.5}
& 41.1\extdrop{$-$1.7} & 38.7\extdrop{$-$3.0} & \underline{49.8}\extgain{+0.2} & \underline{28.4}\extgain{+3.1}
& \textbf{68.8}\extgain{+0.1} & \underline{83.5}\extgain{+2.0} & \underline{93.5}\extgain{+2.5}
& \textbf{38.5}\extgain{+3.7} & \textbf{56.6}\extgain{+11.2} & \textbf{40.8}\extgain{+2.2} \\

\ourmethod
& \textbf{43.6}\extflat{0.0} & \textbf{38.6}\extgain{+2.7}
& \textbf{43.0}\extgain{+0.2} & \textbf{42.3}\extgain{+0.6} & \textbf{50.9}\extgain{+1.3} & \textbf{31.9}\extgain{+6.6}
& \textbf{68.8}\extgain{+0.1} & \textbf{86.5}\extgain{+5.0} & \textbf{94.5}\extgain{+3.5}
& \underline{35.8}\extgain{+1.0} & \underline{48.1}\extgain{+2.7} & \underline{39.5}\extgain{+0.9} \\

\bottomrule
\end{tabular}
\end{adjustbox}
\end{table*}

We evaluate \ourmethod on \ourbenchmark and out-of-domain benchmarks, addressing three questions:
(1) What is missing in current evaluation?
(2) Does our method help?
(3) Does it generalize?

\begin{findingbox}{1}
Existing models struggle to produce both correct reasoning
traces and correct final answers.
\end{findingbox} 
\noindent As shown in \cref{tab:main_benchmark_results}, existing MFMs exhibit a large gap between final-answer accuracy and complete reasoning correctness.
The evaluated video reasoning models achieve MCQ accuracies of 32.6--36.4\%, but their \textbf{Joint} accuracy scores range from only 5.0\% to 8.1\%.
Even Gemini 3.1 Pro, the strongest external baseline in terms of Joint accuracy, achieves 19.1\% Joint accuracy despite an MCQ accuracy of 47.8\%.
These results reveal failures overlooked by answer-only
evaluation, which \ourbenchmark captures by evaluating
individual claims and complete reasoning traces.

\begin{findingbox}{2}
Process rewards improve intermediate-trace correctness beyond final-answer accuracy.
\end{findingbox} 
\noindent As shown in \cref{tab:main_benchmark_results}, \ourmethod achieves the highest \textbf{Exact} and \textbf{Joint} scores among all evaluated models, reaching 24.5\% and 22.9\%, respectively.
Compared with the base model, \ourmethod improves \textbf{Exact} and \textbf{Joint} accuracy by 15.0 and 13.6 percentage points, respectively, and exceeds Gemini 3.1 Pro by 5.4 and 3.8 percentage points.
More importantly, compared with Outcome-only under the same training setting, \ourmethod improves MCQ accuracy by 3.5 percentage points, while also increasing \textbf{Claim Rec.}, \textbf{Exact} accuracy, and \textbf{Joint} accuracy by 2.4, 1.5, and 0.7 percentage points, respectively.
A claim-level analysis shows that the improvement in Claim Recall is primarily driven by visually inferred claims rather than question-visible claims.
To assess trace correctness while controlling for final-answer accuracy, we examine the 434 examples answered correctly by both models.
Within this subset,  \ourmethod improves exact-trace accuracy
from 47.9\% (208) to 52.8\% (229), correcting 24 traces and
introducing 3 errors (two-sided exact McNemar, $p<0.001$).

\begin{figure}[t]
    \centering
    \includegraphics[width=\columnwidth]{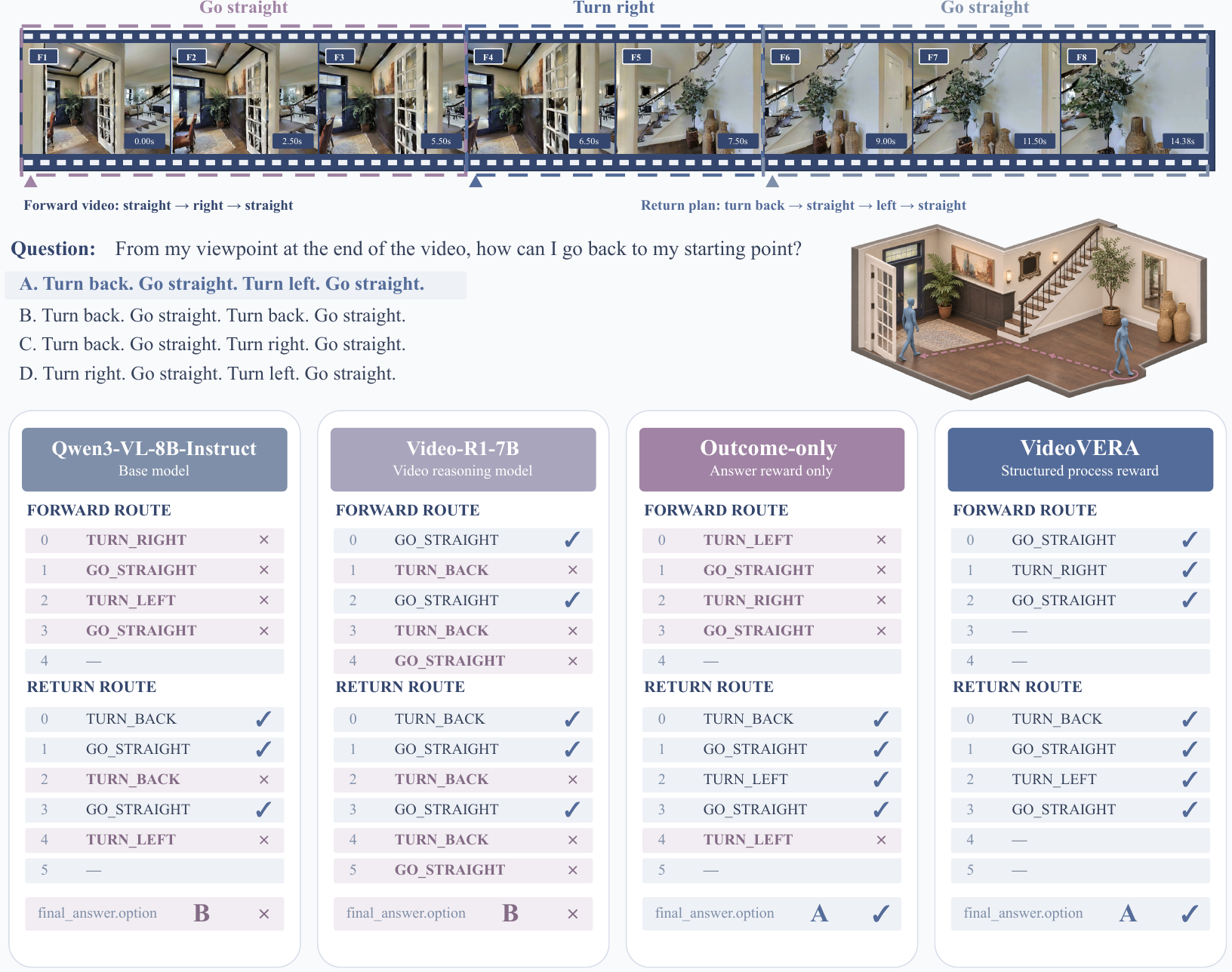}

    \caption{
    \textbf{Qualitative error analysis on reverse route planning.}
    The base model and Video-R1 produce incorrect answers.
    Outcome-only selects the correct answer despite route errors,
    whereas \ourmethod correctly reconstructs both routes.
    }

    \label{fig:four_model_comparison}
\end{figure}

As illustrated in \cref{fig:four_model_comparison}, Outcome-only selects the correct answer despite incorrect route steps, whereas \ourmethod correctly predicts both the answer and route traces.

\begin{findingbox}{3}
Process supervision generalizes to out-of-domain video benchmarks.
\end{findingbox}
\noindent As shown in \cref{tab:external_benchmark_results},
\ourmethod improves overall accuracy on SAW-Bench
by 1.9 percentage points compared with Outcome-only,
with gains in direction (+3.6 pp), shape (+1.1 pp),
and reverse planning (+3.5 pp).
It also improves relative direction on ReVSI (+1.2 pp), action antonym on MVBench (+3.0 pp), and scene transition
(+1.0 pp), while maintaining their overall scores.
However, \ourmethod underperforms Outcome-only on TOMATO by 2.7 pp overall.
Among 79 TOMATO human-demonstration questions with non-segmented reference directions, \ourmethod answers 17 correctly, compared with 33 for Outcome-only.
We observe cases of over-segmented motion prediction (e.g., ``right, then left'' instead of ``left'').
These errors may reflect a transfer limitation between
camera-centric displacement reasoning and TOMATO's hand- or object-centric motion understanding.
Overall, these results show that transfer benefits
vary across tasks.

\section{Analysis}
\label{sec:analysis}
\begin{figure}[t]
    \centering
    \includegraphics[width=0.8\columnwidth]{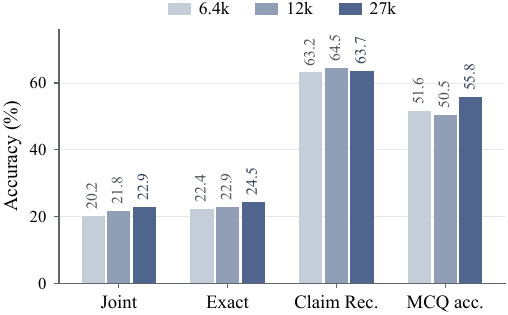}
    \caption{
    \textbf{Effect of training data scale.}
    Exact and Joint accuracy increase across the 6.4k, 12k, and 27k settings, while Claim Rec. remains stable and MCQ accuracy varies across scales.    
    }
    \label{fig:data_scale}
\end{figure}

\subsection{Effect of Training Data Scale.}
As shown in \cref{fig:data_scale}, we evaluate \ourmethod
at three training scales: 6.4k, 12k, and 27k examples.
\textbf{Exact} accuracy increases from 22.4\% to 24.5\%
and \textbf{Joint} accuracy from 20.2\% to 22.9\%
across these scales.
In contrast, \textbf{Claim Rec.} remains relatively stable
(63.2--64.5\%), while MCQ accuracy varies from 51.6\%
at 6.4k to 50.5\% at 12k before reaching 55.8\% at 27k.
These results suggest that increasing training data may benefit complete trace correctness more consistently than individual claim matching.

\subsection{Ablation Study on Reward Components.}
\label{sec:reward_ablation}

\begin{table}[t]
\centering

\caption{
\textbf{Reward ablation on \ourbenchmark{} (6.4k).}
All variants include format and answer rewards; ``+'' denotes added components.
Binary and graded process rewards use exact and partial trace matching, respectively.
Weights are fixed without renormalization.
\textbf{Bold}: best; \underline{underline}: second best.
}

\label{tab:reward_ablation}

\scriptsize
\setlength{\tabcolsep}{2pt}
\renewcommand{\arraystretch}{1.18}

\begin{tabularx}{\columnwidth}{
@{}l*{4}{>{\centering\arraybackslash}X}@{}
}
\toprule
\textbf{Reward Variant}
& \textbf{Avg.}
& \mbox{\textbf{Claim Rec.}}
& \textbf{Exact}
& \textbf{Joint} \\
\midrule

Format + Answer
& \underline{51.2}
& 58.5
& 19.8
& 18.5 \\

+ Graded Process
& 50.9
& \textbf{64.0}
& \underline{21.2}
& \underline{19.5} \\

+ Binary Process + Joint
& 50.1
& 59.7
& 20.0
& 18.7 \\

\midrule
\textbf{+ Graded Process + Joint (Ours)}
& \textbf{51.6}
& \underline{63.2}
& \textbf{22.4}
& \textbf{20.2} \\

\bottomrule
\end{tabularx}
\end{table}

We examine process and joint rewards using the 6.4k
training setting (\cref{tab:reward_ablation}). All variants retain format and answer rewards.

\noindent\textbf{Does process supervision improve trace correctness?}
Adding graded process rewards increases \textbf{Claim Rec.}
from 58.5\% to 64.0\%, \textbf{Exact} from 19.8\% to
21.2\%, and \textbf{Joint} from 18.5\% to 19.5\%.
Including the joint reward further improves \textbf{Exact}
and \textbf{Joint} to 22.4\% and 20.2\%, respectively, with the highest MCQ accuracy (51.6\%).

\noindent\textbf{Is graded feedback more effective than binary supervision?}
Replacing graded process rewards with binary exact-trace
rewards reduces \textbf{Claim Rec.} from 63.2\% to 59.7\%,
\textbf{Exact} from 22.4\% to 20.0\%, and \textbf{Joint} from 20.2\% to 18.7\%.
This suggests that graded rewards provide more useful
feedback than binary rewards for fully correct traces.

\subsection{Robustness and Visual Evidence Dependence}
\label{sec:robustness}
\begin{figure}[t]
    \centering
    \includegraphics[width=\columnwidth]{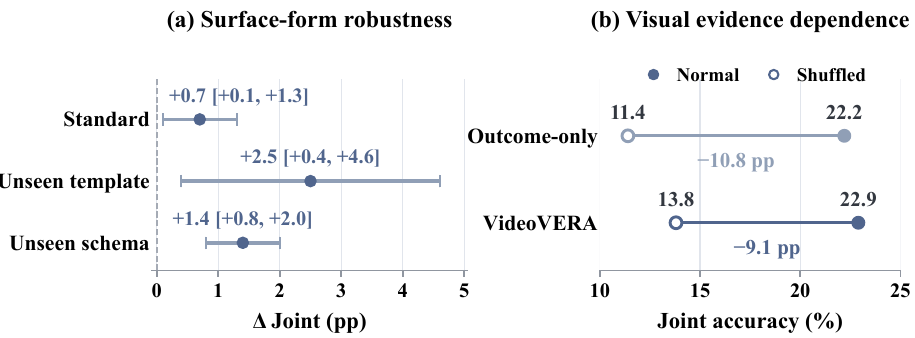}
    \caption{
    \textbf{Robustness analysis.}
    (a) Joint gains over Outcome-only with 95\% paired bootstrap
    confidence intervals.
    (b) Joint accuracy with normal and same-task shuffled videos.
    }
    \label{fig:robustness}
\end{figure}

A potential concern is that \ourmethod may learn task-specific question templates or output schemas rather than transferable video reasoning.
To examine this, we evaluate the same 1,000 benchmark
examples with unseen question templates (rephrased
questions) and unseen output schemas (changed field
names and output structures), while preserving the
underlying reasoning targets.
Predictions under unseen schemas are deterministically mapped back to the original representation for evaluation.
As shown in \cref{fig:robustness}(a), \ourmethod maintains higher \textbf{Joint} accuracy than Outcome-only across all three settings, with gains of 0.7, 2.5, and 1.4 percentage points, respectively.
All three gains have 95\% paired bootstrap confidence intervals excluding zero, suggesting that the benefits of process supervision cannot be explained solely by memorizing training templates or output schemas.

We further examine whether successful reasoning depends on matched visual evidence by replacing each video with another from the same task while keeping the question, answer choices, and reference trace unchanged.
As shown in \cref{fig:robustness}(b), \textbf{Joint} accuracy drops from 22.2\% to 11.4\% for Outcome-only and from 22.9\% to 13.8\% for \ourmethod.
This degradation indicates that answer-and-trace predictions depend on the corresponding video rather than solely on textual or task-specific priors.

\paragraph{Can the answer reconstruct the trace?}
We regenerate traces with and without video, conditioning on each model's original predicted answer.
We first evaluate examples originally answered correctly by each model.
Removing the video reduces \textbf{Exact} accuracy from 41.9\% to 35.6\% for Outcome-only and from 44.4\% to 36.4\% for \ourmethod, with the 95\% confidence intervals for both paired decreases excluding zero.
In \ourbenchmark, 150 route-action claims from 100 Route-Shape examples have identical questions and correct answers but different gold actions across videos. With video, \ourmethod matches 45/150 claims (30.0\%); without video, model-based and rule-based reconstructions conditioned on questions and original predicted answers each match 0/150 (0.0\%).
On the 38 claims from examples where \ourmethod's original answer was correct, its video-conditioned traces match 24 reference claims, while both reconstructions without video match none. The paired examples show that the question and correct answer do not uniquely determine these route actions.

\section{Conclusion}
We introduced \ourmethod, a scalable pipeline that converts video annotations into questions, answers, and reference reasoning traces for deterministic verification.
With this pipeline, we constructed \ourdata and \ourbenchmark across eight spatial, temporal, and route reasoning tasks, enabling process supervision and evaluation beyond final-answer accuracy.
Training with verifiable process rewards improves both answer accuracy and reasoning-trace correctness over Outcome-only.
Across training scales, exact-trace and joint accuracy increase consistently, while reward ablations support graded process feedback.
Our work demonstrates that existing video annotations provide a scalable source of verifiable process supervision and reproducible evaluation for video reasoning.
\section{LLM Usage Statement}
We used ChatGPT to assist with language editing, literature search, code debugging, and experimental analysis. The research ideas, methodology, experimental design, and conclusions were conceived and determined solely by the authors, who take full responsibility for all content. No LLMs were used for dataset generation or annotation.

{
  \small
  \renewcommand{\bibsection}{\section*{References}}
  \bibliographystyle{ieeenat_fullname}
  \bibliography{main}
}

\clearpage
\onecolumn
\pagestyle{supp}
\titleformat{\section}{\bfseries\fontsize{12.5}{15}\selectfont\color{AccentPurple}}
  {\thesection}{0.8em}{}

{\bfseries\fontsize{12.5}{15}\selectfont\color{AccentPurple}
\phantomsection\addcontentsline{toc}{section}{Supplementary Material}
Supplementary Material\par}
\vspace{4pt}

\begin{tcolorbox}[enhanced,colback=AccentPurpleLight,colframe=AccentPurpleLight,
  boxrule=0pt,arc=0pt,borderline west={2.2pt}{0pt}{AccentPurple},
  left=12pt,right=10pt,top=6pt,bottom=6pt,before skip=2pt,after skip=10pt]
{\bfseries\small\color{AccentPurple}Contents}\par\vspace{3pt}
\begingroup
\small
\makeatletter
\newcommand{\suppoutlineentry}[2]{%
  \par\begingroup
  \let\texorpdfstring\@secondoftwo
  \leftskip=\dimexpr#1\dimexpr1em\relax\relax
  \rightskip=1.6em
  \parfillskip=-\rightskip
  \parindent=0pt
  \ifnum#1=0\relax\bfseries\fi
  \hyperref[#2]{%
    \makebox[2.4em][l]{\ref*{#2}}\nameref*{#2}%
    \nobreak\leaders\hbox{\normalfont\kern0.25em.\kern0.25em}\hfill
    \nobreak\makebox[1.6em][r]{\pageref*{#2}}}%
  \par\endgroup
}
\makeatother
\setlength{\baselineskip}{13pt}
\noindent\begin{minipage}[t]{0.48\linewidth}
\suppoutlineentry{0}{sec:supp_data_construction}
\suppoutlineentry{1}{sec:supp_data_partitions}
\suppoutlineentry{1}{sec:supp_generation_rules}
\suppoutlineentry{1}{sec:supp_split_isolation}
\suppoutlineentry{1}{sec:supp_real_examples}
\suppoutlineentry{0}{sec:supp_trace_verification}
\suppoutlineentry{1}{sec:supp_output_contract}
\suppoutlineentry{1}{sec:supp_verifier_metrics}
\end{minipage}\hfill
\begin{minipage}[t]{0.48\linewidth}
\suppoutlineentry{0}{sec:supp_human_quality}
\suppoutlineentry{0}{sec:supp_training_evaluation}
\suppoutlineentry{0}{sec:supp_process_analyses}
\suppoutlineentry{1}{sec:supp_claim_provenance}
\suppoutlineentry{1}{sec:supp_paired_trace}
\suppoutlineentry{1}{sec:supp_answer_reconstruction}
\suppoutlineentry{0}{sec:supp_external_results}
\end{minipage}\par
\endgroup

\end{tcolorbox}

\begin{table*}[!b]
\centering
\caption{
\textbf{Dataset composition and partitions.}
T, C, and R denote training examples, cold-start examples,
and RL prompts, with T = C + R.
B denotes benchmark examples.
EPIC-100 abbreviates EPIC-KITCHENS-100.
R2R/RxR videos are rendered in Matterport3D.
}

\label{tab:supp_composition}
\scriptsize
\setlength{\tabcolsep}{3pt}
\renewcommand{\arraystretch}{1.10}
\begin{tabular*}{\textwidth}{
@{\extracolsep{\fill}}llrrrrrrrrrr@{}
}
\toprule
& & \multicolumn{3}{c}{6.4k}
& \multicolumn{3}{c}{12k}
& \multicolumn{3}{c}{27k} & \\
\cmidrule(lr){3-5}
\cmidrule(lr){6-8}
\cmidrule(lr){9-11}
Task & Annotation sources
& T & C & R
& T & C & R
& T & C & R
& B \\
\midrule
Object Direction
& ScanNet/ScanRefer, ARKitScenes
& 900 & 110 & 790
& 1,800 & 220 & 1,580
& 1,947 & 220 & 1,727
& 125 \\

Egocentric Displacement
& R2R/RxR
& 900 & 110 & 790
& 1,800 & 220 & 1,580
& 4,200 & 220 & 3,980
& 125 \\

Forward Route Planning
& R2R/RxR
& 1,000 & 140 & 860
& 1,631 & 280 & 1,351
& 4,700 & 280 & 4,420
& 125 \\

Landmark Route Planning
& ARKitScenes
& 500 & 60 & 440
& 1,000 & 120 & 880
& 2,109 & 120 & 1,989
& 125 \\

Route-Shape Classification
& R2R/RxR
& 1,000 & 120 & 880
& 2,000 & 240 & 1,760
& 4,700 & 240 & 4,460
& 125 \\

Reverse Route Planning
& R2R/RxR
& 1,000 & 120 & 880
& 2,000 & 240 & 1,760
& 4,624 & 240 & 4,384
& 125 \\

Event-Sequence Ordering
& EPIC-100
& 500 & 60 & 440
& 1,000 & 120 & 880
& 2,149 & 120 & 2,029
& 125 \\

Adjacent-Action Identification
& EPIC-100
& 600 & 80 & 520
& 1,200 & 160 & 1,040
& 2,800 & 160 & 2,640
& 125 \\
\midrule
Total
&
& 6,400 & 800 & 5,600
& 12,431 & 1,600 & 10,831
& 27,229 & 1,600 & 25,629
& 1,000 \\
\bottomrule
\end{tabular*}
\end{table*}

\appendix
\renewcommand{\thesection}{\Alph{section}}

\definecolor{ExampleHeader}{HTML}{D8DEE9}
\definecolor{ExampleSpatial}{HTML}{EDF0F5}
\definecolor{ExampleRoute}{HTML}{F0EDF3}
\definecolor{ExampleTemporal}{HTML}{EDF1ED}

\begin{table*}[!b]
\centering
\caption{
\textbf{Task-specific generation rules.}
The table describes how source annotations determine answers,
reference claims, and candidate selection.
}
\label{tab:supp_generation_rules}
\scriptsize
\setlength{\tabcolsep}{4pt}
\renewcommand{\arraystretch}{1.15}
\begin{tabularx}{\textwidth}{
@{}
>{\raggedright\arraybackslash}p{0.13\textwidth}
>{\raggedright\arraybackslash}p{0.19\textwidth}
>{\raggedright\arraybackslash}X
>{\raggedright\arraybackslash}p{0.24\textwidth}
@{}
}
\toprule
Task
& Source annotations
& Answer and reference construction
& Main selection rules \\
\midrule

Object Direction
& Object identities, centers, extents, and visibility.
& Project the target offset onto observer--facing axes.
Record object roles, front/back, left/right, and the
combined quadrant.
& Reject ambiguous geometry, target extents crossing
quadrant boundaries, and insufficient object visibility. \\

Egocentric Displacement
& Camera positions, rotations, timestamps, and endpoints.
& Express endpoint displacement in the reference camera
frame. Record endpoints, movement/turn claims, and direction.
& Reject ambiguous sector boundaries, unstable camera
axes, invalid poses, and reused routes. \\

Forward Route Planning
& Rendered camera trajectories and pose-bound landmarks.
& Extract straight legs and turns. Blank turns in the
question and retain the complete indexed action sequence.
& Require sufficiently long legs, stable headings,
unambiguous turns, and distinct landmark descriptions. \\

Landmark Route Planning
& Object boxes, RGB/depth, intrinsics, and camera poses.
& Compute turns between object-center route legs.
Record the start, facing landmark, turns, and destinations.
& Require visible landmarks, continuous video, valid
depth support, and bounded route-leg distances. \\

Route-Shape Classification
& Traveled-leg headings, turn roles, and stationary ranges.
& Classify traveled-path bends as straight, L, U,
or Manhattan. Classify stationary rotation as in-place
orientation. Retain actions and the shape conclusion.
& Separate path bends from viewing rotations.
Recompute labels after cropping and limit windows per parent. \\

Reverse Route Planning
& Forward legs, actions, and final camera heading.
& Starting from the final heading, turn toward and
traverse each leg in reverse order. Record forward
and reverse action sequences.
& Recheck answers after rendering or cropping
and reject repeated media. \\

Event-Sequence Ordering
& Action identities, labels, and temporal boundaries.
& Sort four distinct actions chronologically.
Use alternative permutations as distractors and
record four ordered-event claims.
& Reject excessive overlap and require sampled-frame
support for every action. \\

Adjacent-Action Identification
& Complete annotation order, anchor, and query direction.
& Select the immediate predecessor or successor.
Record event observations and the requested relation.
& Reject missing neighbors, repeated action identities,
excessive overlap, and insufficient sampled support. \\

\bottomrule
\end{tabularx}
\end{table*}

\begin{table*}[!b]
\centering
\caption{
\textbf{Annotation-to-trace examples.}
Three records from the 27k RL partition.
Trace notation is abbreviated for readability.
}
\label{tab:supp_generation_examples}

\small
\setlength{\tabcolsep}{5pt}
\renewcommand{\arraystretch}{1.15}

\begin{tabularx}{\textwidth}{
>{\raggedright\arraybackslash}p{0.08\textwidth}
>{\raggedright\arraybackslash}X
>{\raggedright\arraybackslash}p{0.27\textwidth}
>{\raggedright\arraybackslash}p{0.25\textwidth}
}
\toprule
\textbf{Family}
& \textbf{Question summary}
& \textbf{Source measurements}
& \textbf{Reference trace and answer} \\
\midrule

\rowcolor{ExampleSpatial}
\textbf{Spatial}
& Stand at the bed, face the red couch, and determine
the TV's relative direction.
& Target offsets: $-0.6246$\,m forward and
$-2.0314$\,m right, giving back-left.
& Observer: bed; facing: couch; target: TV;
longitudinal: back; lateral: left;
direction: back-left.
\newline \textbf{Answer: back-left.} \\

\addlinespace[3pt]

\rowcolor{ExampleRoute}
\textbf{Route}
& Fill in the turn between two specified straight legs.
Options: turn back, turn right, or turn left.
& Heading change: $-68.46^\circ$, giving a right turn.
& Go straight $\rightarrow$ turn right
$\rightarrow$ go straight.
\newline \textbf{Answer: B.} \\

\addlinespace[3pt]

\rowcolor{ExampleTemporal}
\textbf{Temporal}
& Order four actions. Distractors permute the same actions.
& Start times: take pot (3.10\,s), put pot (4.10\,s),
take spoon (6.58\,s), stir oats (7.80\,s).
& Take pot $\rightarrow$ put pot $\rightarrow$
take spoon $\rightarrow$ stir oats, indexed 0--3. \\

\bottomrule
\end{tabularx}
\end{table*}

\section{Data Construction and Dataset Statistics}
\label{sec:supp_data_construction}
We construct each example from source annotations, producing a continuous video, a multiple-choice question, a correct option, and an ordered reference trace.
Task-specific rules convert geometric or temporal annotations into the claims required by the output schema.

\subsection{Dataset Composition and Partitions}
\label{sec:supp_data_partitions}
Table~\ref{tab:supp_composition} summarizes the three training scales and the benchmark.
Cold start uses 800 examples at 6.4k and the main paper's
1,600-example configuration at 12k/27k.
The remaining examples serve as RL prompts.
Full-data SFT uses both partitions.
The 27k release includes 12,431 examples from the 12k
release and uses its cold-start partition.
Within each scale, cold-start and RL example identifiers
are disjoint.

Training task proportions reflect the availability of candidates passing source-specific geometry, visibility, deduplication, and scene constraints.
We train without task resampling, using the same ordered RL prompt pool for all controlled variants at each scale.
The benchmark contains 125 examples per task, so average
MCQ accuracy weights all eight tasks equally.

\subsection{Task-Specific Generation Rules}
\label{sec:supp_generation_rules}
Table~\ref{tab:supp_generation_rules} summarizes how source annotations determine answers and reference claims.
Questions specify the objects, endpoints, landmarks,
or temporal anchors.
We construct distractors from alternative direction labels, action sequences, or event orders, then permute the options.

\paragraph{Spatial computations.}
For Object Direction, let $o$, $f$, and $t$ denote the
horizontal centers of the observer, facing landmark, and target.
We define forward and right axes as
\[
\mathbf{u}=\frac{f-o}{\|f-o\|},
\qquad
\mathbf{r}=(u_y,-u_x).
\]
The signs of $(t-o)^\top\mathbf{u}$ and $(t-o)^\top\mathbf{r}$ determine the target quadrant.
Both component magnitudes must reach 0.25\,m, and the
projected target extent must remain at least 0.10\,m
from either quadrant boundary.
For Egocentric Displacement, we derive forward and right
axes from the reference camera pose.
Distances below 0.10\,m indicate the same location,
distances from 0.10 to 0.30\,m are rejected, and larger
displacements use eight 45-degree sectors.
Bearings within 5 degrees of a sector boundary are rejected.

\paragraph{Route computations.}
For Forward Route Planning, we simplify measured paths
with a 0.20\,m deviation tolerance and require traveled
legs of at least 0.60\,m.
Heading changes below 35 degrees indicate no turn,
changes of at least 150 degrees indicate a back turn,
and intermediate positive/negative changes indicate left/right turns.
We reject changes within 5 degrees of either threshold.
For Landmark Route Planning, object-center legs span
0.70--5.5\,m, with left/right turns of 50--125 degrees
and back turns of at least 150 degrees.
For route shapes, bends span 60--120 degrees.
One bend gives an L-shape, while two same-sign bends
totaling 150--210 degrees give a U-shape.
Other accepted multi-bend routes receive the Manhattan label.
In-place orientation requires at most 0.01\,m displacement and at least 60 degrees of net rotation.
Reverse plans account for the final camera heading before
retracing the forward legs.
Forward-route landmark descriptions use temperature-zero
Qwen inference with confidence and uniqueness checks;
measured trajectories determine the action targets.

\paragraph{Temporal computations.}
We order EPIC-KITCHENS-100 segments by start frame,
start time, and segment identifier.
Event-ordering windows contain four distinct actions
within 18\,s.
Adjacent-action pairs come from consecutive annotations
in the full recording and span at most 8\,s.
Both tasks reject adjacent segments whose overlap exceeds
55\% of the shorter duration.
Each required action must appear in at least two of the
24 uniformly sampled model frames.

\subsection{Retention and Split Isolation}
\label{sec:supp_split_isolation}

We retain examples with decodable videos, distinct answer
options, schema-valid traces, and reference responses
receiving full reward.
Task-specific visibility and trajectory checks establish
visual support before admission.
Duplicate QA slots, defined by task, sampled-video signature,
and normalized question text, are rejected.

Training and benchmark examples are separated by identifier,
source group, and sampled decoded content.
Source groups correspond to physical ScanNet scenes,
Matterport3D scenes, ARKitScenes visits, or EPIC recordings.
For content checks, we sample 24 uniformly spaced decoded
frames, resize each to $64\times64$ RGB, and hash the
concatenated pixels.
The 27k audit reports no benchmark source-group overlap,
matching content signatures, or duplicate QA slots.
Within training, distinct questions may share source media.

\subsection{Annotation-to-Trace Examples}
\label{sec:supp_real_examples}
Table~\ref{tab:supp_generation_examples} presents three examples
from the 27k RL partition.
We abbreviate the trace notation for readability.

\section{Trace Format and Verification}
\label{sec:supp_trace_verification}
\begin{table*}[!b]
\centering
\caption{
\textbf{Reference trace structure across tasks.}
Arrows indicate reference order.
Max. steps gives the output limit.
}
\label{tab:supp_trace_contract}

\footnotesize
\setlength{\tabcolsep}{5pt}
\renewcommand{\arraystretch}{1.10}

\begin{tabularx}{\textwidth}{
@{}
>{\raggedright\arraybackslash}p{0.23\textwidth}
>{\raggedright\arraybackslash}X
c
@{}
}
\toprule
\textbf{Task}
& \textbf{Reference reasoning steps}
& \textbf{Max. steps} \\
\midrule

Object Direction
& Observer object $\rightarrow$ facing object
$\rightarrow$ target object $\rightarrow$ front/back
$\rightarrow$ left/right $\rightarrow$ final direction
& 6 \\

\addlinespace[1pt]

Egocentric Displacement
& Reference and target endpoints $\rightarrow$
movements and turns $\rightarrow$ final direction
& 9 \\

\addlinespace[1pt]

Forward Route Planning
& Complete sequence of straight movements and turns
& 9 \\

\addlinespace[1pt]

Landmark Route Planning
& Starting landmark $\rightarrow$ initial orientation
$\rightarrow$ turns and destination landmarks
$\rightarrow$ full turn sequence
& 9 \\

\addlinespace[1pt]

Route-Shape Classification
& Movements and turns $\rightarrow$
stationarity for in-place rotation $\rightarrow$ route shape
& 7 \\

\addlinespace[1pt]

Reverse Route Planning
& Forward actions $\rightarrow$ reverse actions
$\rightarrow$ full reverse plan
& 12 \\

\addlinespace[1pt]

Event-Sequence Ordering
& Four observed actions in chronological order
& 4 \\

\addlinespace[1pt]

Adjacent-Action Identification
& Two observed actions in chronological order
$\rightarrow$ before/after relation
& 3 \\

\bottomrule
\end{tabularx}
\end{table*}

\subsection{Output Contract}
\label{sec:supp_output_contract}
Each response contains a task identifier, a sequence of claims, and a final answer in JSON format.
A claim slot is identified by its type and applicable index, phase, or landmark role.
Reward validation requires schema validity, consecutive step identifiers, unique claim slots, and an option letter present in the question.
The verifier lowercases object and action labels,
trims surrounding whitespace and trailing periods,
and collapses internal whitespace.
Categorical values remain unchanged, and no synonym matching is used.
Table~\ref{tab:supp_trace_contract} summarizes the
task-specific claim types, reference order, and trace-length limits.

\subsection{Rewards and Evaluation Metrics}
\label{sec:supp_verifier_metrics}

\paragraph{Ordered process reward.}
Let $G$ and $\hat{G}$ contain $M$ reference claims and
$N$ predicted claims, respectively.
For a reward-valid response, let $K$ be the length of their
longest common subsequence under exact normalized claim matching.
The process reward is
\begin{equation}
R_p=\frac{2K}{M+N}.
\label{eq:supp_process_reward}
\end{equation}
An empty trace receives $R_p=0$.
Missing claims do not prevent later claims from receiving credit.
Surplus claims increase $N$ without adding matches,
while incorrect claim values can reduce $K$.
With reward-validity indicator $V$, answer-correctness indicator $A$,
and ordered trace agreement
$E_{\mathrm{ord}}=\mathbf{1}[\hat{G}=G]$,
the structured reward is
\begin{equation}
R=V\left(
0.05+0.40A+0.45R_p+0.10AE_{\mathrm{ord}}
\right).
\label{eq:supp_structured_reward}
\end{equation}
Outcome-only uses $VA$.
A wrong answer removes the answer and joint components
but preserves valid process credit.

\paragraph{Reported metrics.}
For example $i$, let $c_i$ count reference slots with an
exactly matching predicted claim and $M_i$ count reference claims.
Claim recall is
\begin{equation}
\mathrm{Claim\ Rec.}
=\frac{\sum_i c_i}{\sum_i M_i}.
\label{eq:supp_claim_recall}
\end{equation}
Unparseable or schema-invalid responses receive zero claim credit.
For schema-valid responses, additional or duplicate claims
do not increase the reference-claim denominator.
Exact requires a schema-valid response with consecutive step
identifiers and all reference claims correctly predicted in
reference order, with no missing, additional, or duplicated claims.
Joint additionally requires a correct final answer.
Duplicate slots invalidate the training reward but may retain
diagnostic claim recall if the response remains schema-valid.

\begin{table}[!b]
\centering
\caption{
\textbf{Human review results.}
Ans., Trace, and Joint denote answer agreement,
trace validity, and joint acceptance (\%).
}
\label{tab:supp_human_review}
\scriptsize
\setlength{\tabcolsep}{3pt}
\renewcommand{\arraystretch}{1.10}

\begin{tabular*}{\columnwidth}{
@{\extracolsep{\fill}}lrrr@{}
}
\toprule
\textbf{Task}
& \textbf{Ans.}
& \textbf{Trace}
& \textbf{Joint} \\
\midrule
OD  & 98.0 & 98.0 & 98.0 \\
ED  & 97.0 & 96.0 & 96.0 \\
FRP & 97.0 & 97.0 & 97.0 \\
LRP & 97.0 & 96.0 & 96.0 \\
RSC & 99.0 & 98.0 & 98.0 \\
RRP & 98.0 & 97.0 & 97.0 \\
ESO & 99.0 & 99.0 & 99.0 \\
AA & 98.0 & 98.0 & 98.0 \\
\midrule
Overall & 97.9 & 97.4 & 97.4 \\
\bottomrule
\end{tabular*}
\end{table}

\section{Human Quality Control}
\label{sec:supp_human_quality}

We sample 800 examples across eight tasks:
400 from training and 400 from the benchmark.
Two reviewers independently assess video answerability and
record answers and reasoning steps without seeing generated targets.
We compare their records with the generated answers and traces,
and resolve disagreements through discussion between the two reviewers.
Unresolved disagreements are referred to a third reviewer for adjudication.
Table~\ref{tab:supp_human_review} reports the results
after adjudication.
Pre-adjudication joint-acceptance agreement yields
Cohen's $\kappa=0.881$.

\section{Training Configuration}
\label{sec:supp_training_evaluation}

Table~\ref{tab:supp_training_configs} summarizes the
SFT and RL hyperparameters.
Training partitions are reported in
Table~\ref{tab:supp_composition}, and evaluation protocols
follow \cref{sec:evaluation_protocol}.

\begin{table}[!b]
\centering
\caption{
\textbf{Training hyperparameters.}
SFT includes full-data and cold-start training.
RL includes Outcome-only and \ourmethod.
}
\label{tab:supp_training_configs}
\scriptsize
\setlength{\tabcolsep}{4pt}
\renewcommand{\arraystretch}{1.12}

\begin{tabular*}{\columnwidth}{
@{\extracolsep{\fill}}lcc@{}
}
\toprule
\textbf{Setting}
& \textbf{SFT}
& \textbf{RL} \\
\midrule
Training passes & 1 & 1 \\
Learning rate & $10^{-4}$ & $5\times10^{-6}$ \\
Learning-rate warm-up steps & -- & 50 \\
LoRA rank & 16 & 16 \\
LoRA scaling & 32 & 32 \\
LoRA dropout & 0.05 & 0 \\
Responses per prompt & -- & 8 \\
Policy epochs per informative group & -- & 2 \\
Rollout temperature & -- & 0.8 \\
Rollout top-$p$ & -- & 0.95 \\
Maximum rollout tokens & -- & 640 \\
GRPO clipping & -- & 0.2 \\
KL coefficient & -- & 0.02 \\
\bottomrule
\end{tabular*}
\end{table}

\section{Additional Process-Level Analyses}
\label{sec:supp_process_analyses}

\subsection{Claim-Level Improvements}
\label{sec:supp_claim_provenance}

We distinguish question-visible claims from visually
inferred claims.
Table~\ref{tab:supp_claim_provenance} reports recall
over the 7,200 reference claims.
Visually inferred claims improve by 3.5 percentage points,
compared with 0.6 for question-visible claims,
accounting for most of the overall improvement.

\begin{table}[!b]
\centering
\caption{
\textbf{Claim recall by information source.}
Overall recall is weighted by reference-claim counts.
Recall and differences are percentages and
percentage points, respectively.
}
\label{tab:supp_claim_provenance}
\scriptsize
\setlength{\tabcolsep}{3pt}
\renewcommand{\arraystretch}{1.10}

\begin{tabular*}{\columnwidth}{
@{\extracolsep{\fill}}lrrrr@{}
}
\toprule
\textbf{Claims}
& \textbf{Count}
& \textbf{Outcome-only}
& \textbf{\ourmethod}
& $\Delta$ \\
\midrule
Question-visible
& 2,800 & 88.0 & 88.6 & +0.6 \\
Visually inferred
& 4,400 & 44.3 & 47.8 & +3.5 \\
\midrule
Overall
& 7,200 & 61.3 & 63.7 & +2.4 \\
\bottomrule
\end{tabular*}
\end{table}

\subsection{Paired Trace Correctness}
\label{sec:supp_paired_trace}

We compare the 434 benchmark examples answered
correctly by both models.
As shown in Table~\ref{tab:supp_trace_transitions}(a),
\ourmethod repairs 24 trace errors while introducing three,
increasing Exact from 208/434 (47.9\%) to
229/434 (52.8\%).
The difference is significant under a two-sided exact
McNemar test ($p<0.001$), holding answer correctness fixed.

\begin{table}[!b]
\centering
\caption{
\textbf{Paired trace correctness and reconstruction.}
(a) Examples answered correctly by both models.
(b) Examples originally answered correctly by Outcome-only;
rows indicate regenerated with-video traces and columns
indicate regenerated text-only traces.
Entries are example counts.
}
\label{tab:supp_trace_transitions}
\small
\setlength{\tabcolsep}{4pt}
\renewcommand{\arraystretch}{1.10}

\begin{tabular*}{\columnwidth}{
@{\extracolsep{\fill}}lrr@{}
}
\toprule
\multicolumn{3}{@{}l}{
\textbf{(a) Cross-model comparison ($N=434$)}
} \\
\addlinespace[3pt]
& \multicolumn{2}{c}{\textbf{\ourmethod trace}} \\
\cmidrule(lr){2-3}
\textbf{Outcome-only trace}
& \textbf{Non-exact}
& \textbf{Exact} \\
\midrule
Non-exact & 202 & 24 \\
Exact     & 3   & 205 \\

\midrule
\multicolumn{3}{@{}l}{
\textbf{(b) Outcome-only reconstruction ($N=523$)}
} \\
\addlinespace[3pt]
& \multicolumn{2}{c}{\textbf{Text-only reconstruction}} \\
\cmidrule(lr){2-3}
\textbf{With-video trace}
& \textbf{Non-exact}
& \textbf{Exact} \\
\midrule
Non-exact & 279 & 25 \\
Exact     & 58  & 161 \\
\bottomrule
\end{tabular*}
\end{table}

\subsection{Answer-Conditioned Reconstruction}
\label{sec:supp_answer_reconstruction}

We regenerate traces with and without video,
conditioning on each model's original predicted answer.
The supplied answer is held fixed in both conditions.
Evaluation uses each model's originally correct-answer
subset, rather than the shared 434-example subset above.
Removing video reduces Exact from 41.9\% to 35.6\%
for Outcome-only and from 44.4\% to 36.4\%
for \ourmethod.

Table~\ref{tab:supp_trace_transitions}(b)
reports the paired Outcome-only results on 523 examples.
With-video and text-only traces achieve Exact on
219/523 and 186/523 examples, respectively.
Without video, 58 previously exact traces become non-exact,
while 25 non-exact traces become exact,
giving a net decrease of 33 examples
(6.31 percentage points).

We additionally examine 150 route-action claims from
100 Route-Shape examples in \ourbenchmark with identical
questions and correct answers but different reference
actions across videos.
With video, \ourmethod matches 45/150 claims (30.0\%);
without video, model-based and rule-based reconstructions
conditioned on questions and original predicted answers
each match none.
Among the 38 claims from examples originally answered
correctly by \ourmethod, video-conditioned traces match
24 claims, while both reconstructions match none.
These examples show that the question and correct answer
do not uniquely determine the reference route actions.

\clearpage
\section{Complete Out-of-Domain Results}
\label{sec:supp_external_results}

Table~\ref{tab:supp_external_complete} expands the external
comparison in Table~\ref{tab:external_benchmark_results}
to all evaluated tasks and categories.
We compare Base with SFT, Outcome-only, and \ourmethod
trained at the 27k scale.
The complete results show task-dependent transfer:
\ourmethod achieves the highest SAW-Bench Overall score
and ties for the highest Overall scores on ReVSI and MVBench,
while Outcome-only performs best overall on TOMATO.
On SAW-Bench, gains in direction, shape, and reverse planning
coexist with lower localization and affordance scores.

\definecolor{ExampleSpatial}{HTML}{EDF0F5}
\definecolor{ExampleRoute}{HTML}{F0EDF3}
\definecolor{ExampleTemporal}{HTML}{EDF1ED}

\begingroup
\scriptsize
\setlength{\tabcolsep}{7pt}
\renewcommand{\arraystretch}{1.03}
\setlength{\LTleft}{0pt}\setlength{\LTright}{0pt}\setlength{\LTcapwidth}{\textwidth}
\begin{longtable}{@{\extracolsep{\fill}}lrrrrr@{}}
\caption{
\textbf{Complete results on external video understanding benchmarks.}
All three trained models use the 27k training scale.
Scores are percentages; $N$ denotes the number of evaluated questions.
ReVSI Overall averages its seven category scores, using the
benchmark's numerical-question metric where applicable.
Other Overall scores are question-weighted averages,
equivalently the task average for MVBench.
SAW-Bench uses the original labels.
\textbf{Bold} and \underline{underline} indicate the best and
second-best distinct displayed scores within each row;
ties receive the same formatting.
}
\label{tab:supp_external_complete}\\
\toprule
\textbf{Task / Category}
& $N$
& \textbf{Base}
& \textbf{SFT (27k)}
& \textbf{Outcome-only (27k)}
& \textbf{\ourmethod{} (27k)} \\
\midrule
\endfirsthead
\multicolumn{6}{l}{\textit{Table \thetable{} (continued)}}\\
\toprule
\textbf{Task / Category}
& $N$
& \textbf{Base}
& \textbf{SFT (27k)}
& \textbf{Outcome-only (27k)}
& \textbf{\ourmethod{} (27k)} \\
\midrule
\endhead
\bottomrule
\endfoot
\bottomrule
\endlastfoot
\rowcolor{ExampleSpatial}
\multicolumn{6}{l}{\textbf{ReVSI}} \\

Object Counting
& 1,090 & 33.4 & \textbf{37.6} & \underline{35.7} & 33.9 \\
Object Size Estimation
& 1,126 & \textbf{66.0} & 64.5 & 65.6 & \underline{65.7} \\
Absolute Distance
& 1,208 & 41.8 & \underline{44.1} & \textbf{46.7} & 43.4 \\
Relative Distance
& 1,279 & \textbf{46.5} & 42.9 & \underline{45.2} & 45.0 \\
Relative Direction
& 1,332 & 35.9 & \underline{38.4} & 37.4 & \textbf{38.6} \\
Route Planning
& 279 & \textbf{37.6} & 30.5
& \underline{32.3} & \underline{32.3} \\
Room Size Estimation
& 302 & \underline{44.3} & 26.0 & 42.5 & \textbf{46.1} \\
\textbf{Overall}
& 6,616 & \textbf{43.6} & \underline{40.6}
& \textbf{43.6} & \textbf{43.6} \\

\midrule
\rowcolor{ExampleSpatial}
\multicolumn{6}{l}{\textbf{SAW-Bench}} \\
Direction
& 834 & \underline{41.7} & 41.4 & 38.7 & \textbf{42.3} \\
Shape
& 546 & 49.6 & 45.8 & \underline{49.8} & \textbf{50.9} \\
Reverse Planning
& 229 & 25.3 & \underline{28.4}
& \underline{28.4} & \textbf{31.9} \\
Localization
& 200 & \textbf{44.5} & \underline{43.0} & 38.0 & 35.0 \\
Affordance
& 162 & \underline{51.2} & \underline{51.2}
& \textbf{53.1} & 50.0 \\
Memory
& 100 & \textbf{37.0} & \textbf{37.0}
& 30.0 & \underline{36.0} \\
\textbf{Overall}
& 2,071 & \underline{42.8} & 41.8 & 41.1 & \textbf{43.0} \\

\midrule
\rowcolor{ExampleRoute}
\multicolumn{6}{l}{\textbf{MVBench}} \\
Object Interaction
& 200 & \underline{77.5} & \textbf{78.5}
& \underline{77.5} & \underline{77.5} \\
Action Sequence
& 200 & \textbf{80.5} & 78.0 & 79.0 & \underline{79.5} \\
Action Prediction
& 200 & \textbf{67.0} & 62.5 & 65.0 & \underline{66.5} \\
Action Localization
& 200 & 57.0 & \textbf{60.0} & \underline{58.5} & 56.0 \\
Moving Count
& 200 & 69.0 & 69.0 & \underline{69.5} & \textbf{70.5} \\
Fine Grained Pose
& 200 & 73.0 & \textbf{75.0}
& \underline{74.5} & \textbf{75.0} \\
Character Order
& 200 & \textbf{78.5} & \underline{77.5}
& \textbf{78.5} & 76.5 \\
Object Shuffle
& 200 & \underline{40.0} & \textbf{44.0} & 37.0 & 38.0 \\
Egocentric Navigation
& 200 & \textbf{37.5} & \underline{33.0} & 32.0 & 31.5 \\
Moving Direction
& 200 & \textbf{55.0} & 48.5
& \underline{53.5} & \textbf{55.0} \\
Episodic Reasoning
& 200 & \textbf{57.0} & 52.5
& \underline{53.0} & \underline{53.0} \\
Fine Grained Action
& 200 & 48.0 & \underline{49.5}
& \textbf{50.0} & \textbf{50.0} \\
Scene Transition
& 200 & 91.0 & \underline{93.5}
& \underline{93.5} & \textbf{94.5} \\
State Change
& 200 & 75.0 & 73.0 & \underline{76.0} & \textbf{76.5} \\
Moving Attribute
& 200 & \textbf{94.0} & \underline{93.5}
& 93.0 & \underline{93.5} \\
Action Antonym
& 200 & 81.5 & \underline{83.5}
& \underline{83.5} & \textbf{86.5} \\
Unexpected Action
& 200 & \textbf{82.5} & 81.0
& \underline{82.0} & \underline{82.0} \\
Counterfactual Inference
& 200 & 68.5 & 69.0 & \textbf{74.0} & \underline{72.5} \\
Object Existence
& 200 & 86.0 & \underline{87.0} & 86.0 & \textbf{87.5} \\
Action Count
& 200 & 55.0 & \textbf{61.0} & \underline{60.0} & 54.5 \\
\textbf{Overall}
& 4,000 & \underline{68.7} & 68.5
& \textbf{68.8} & \textbf{68.8} \\

\midrule
\rowcolor{ExampleTemporal}
\multicolumn{6}{l}{\textbf{TOMATO}} \\

Count
& 292 & 30.1 & 25.7 & \textbf{37.0} & \underline{34.6} \\
Direction
& 403 & 45.4 & 46.9 & \textbf{56.6} & \underline{48.1} \\
Rotation
& 286 & \textbf{22.0} & \underline{18.2}
& \textbf{22.0} & \underline{18.2} \\
Shape \& Trend
& 223 & 38.6 & 35.9 & \textbf{40.8} & \underline{39.5} \\
Velocity \& Frequency
& 210 & \underline{32.4} & 28.6 & 26.2 & \textbf{32.9} \\
Visual Cues
& 70 & \textbf{41.4} & \textbf{41.4}
& 37.1 & \underline{38.6} \\
\textbf{Overall}
& 1,484 & 34.8 & 32.7 & \textbf{38.5} & \underline{35.8} \\

\end{longtable}
\endgroup

\end{document}